\documentclass[10pt,twocolumn,letterpaper]{article}

\usepackage{wacv}
\usepackage{times}
\usepackage{epsfig}
\usepackage{graphicx}
\usepackage{amsmath}
\usepackage{amssymb}
\usepackage{makecell}
\usepackage{multirow}

\usepackage{times}
\usepackage[latin9]{inputenc}
\usepackage[english]{babel}
\usepackage{caption}% <-- added
\usepackage[colorlinks = true,
            linkcolor = magenta,
            urlcolor  = magenta,
            citecolor = magenta,
            anchorcolor = magenta]{hyperref}
\usepackage{tabulary}
\usepackage{textcomp}

\usepackage[para]{threeparttable}
\usepackage{array,booktabs,longtable,tabularx}
\usepackage{ltablex}% <-- added
\usepackage{siunitx}% <-- added
\usepackage{caption}% <-- added
\usepackage[flushleft]{threeparttablex}
\usepackage{footnote}
\makesavenoteenv{tabular}
\makesavenoteenv{table}

\wacvfinalcopy % *** Uncomment this line for the final submission

\def\wacvPaperID{430} % *** Enter the wacv Paper ID here

\begin{document}

%%%%%%%%% TITLE
\title{Development of Real-time ADAS Object Detector for Deployment on CPU}

% Authors at the same institution
%\author{First Author \hspace{2cm} Second Author \\
% Authors at different institutions
\author{Alexander Kozlov \\
Intel\\
{\tt\small alexander.kozlov@intel.com}
\and
Daniil Osokin \\
Intel\\
{\tt\small daniil.osokin@intel.com}
}

\maketitle
\ifwacvfinal\thispagestyle{empty}\fi

%%%%%%%%% ABSTRACT
\begin{abstract}
   In this work, we outline the set of problems, which any Object Detection CNN faces when its development comes to the deployment stage and propose methods to deal with such difficulties. We show that these practices allow one to get Object Detection network, which can recognize two classes: vehicles and pedestrians and achieves more than 60 frames per second inference speed on Core\textsuperscript{TM} i5-6500 CPU. The proposed model is built on top of the popular Single Shot MultiBox Object Detection framework but with substantial improvements, which were inspired by the discovered problems. The network has just 1.96 GMAC\footnote{GMAC -- billions of multiply-accumulate operations.} complexity and less than 7 MB model size. It is publicly available as a part of Intel\textsuperscript{\textregistered{}} OpenVINO\textsuperscript{TM} Toolkit.
\end{abstract}

%%%%%%%%% BODY TEXT
\section{Introduction}

Object detection (OD) is an important cue in developing products for many domains: Digital Security and Surveillance, Autonomous vehicles, etc. Usually, complex solutions solve multiple tasks in parallel, so it is essential to have a fast and accurate algorithm.

In \cite{SpeedAccTradeoff} authors compared modern meta-architectures for OD and shown, that there is a speed/accuracy trade-off: higher quality two-stage models work slower than less accurate single shot methods. 
However, even single shot methods \cite{SSD} work more or less fast only on small resolution (e.g., 300x300 pixels).

For deployment\footnote{By deployment, we understand the steps that should be done in order to use the algorithm in market-ready solutions.}, it is not enough to have just a fast object detector. It has to be robust against the two typical types of errors: missing objects and false alarms.

In this work, we address the problem of designing an object detector for deployment. Specifically, how to make it fast, yet pretty accurate, as well as compensate its blinking and false positives. We selected autonomous driving domain and require detection of 2 classes of objects: vehicles and pedestrians, which is a common task in advanced driving-assistance systems (ADAS).
Our main contributions are:

\begin{itemize}
    \item We advise the design steps to build an OD, which can run at 60+ fps on edge devices.
    \item We provide a designed model available for evaluation as a part of an open source inference framework.
    \item We show that it is possible to train detector without ImageNet\cite{Imagenet} pre-training.
    %\item We show, how to train detectors without ImageNet\cite{Imagenet} pre-training.
    \item We propose the way to weaken the confidence of typical false positives that helps to reduce false alarm rate.
    \item We present a lightweight strategy to post-process detections in order to compensate their blinking.
\end{itemize}
%-------------------------------------------------------------------------
\subsection{Related Work}

%Huge progress in OD was linked to various detector meta-architectures design.
There are two major groups of DL OD: one and two-stage methods. For two-stage methods, Faster R-CNN \cite{Faster} provides the best quality, but it is the slowest one. R-FCN \cite{Rfcn} aims to improve the speed by making all computations shared with position sensitive score maps but at the cost of accuracy. One-stage methods, such as SSD \cite{SSD} are the fastest ones. However, their speed degrades on high-resolution input.

An important part of research is conducted by the design of lightweight backbones, which can perform on par with the top networks for classification. CNNs, that utilize depth-wise convolutions \cite{MobileNet}, \cite{Xception}, allow achieving dramatic parameter reduction and faster inference time. Authors in \cite{SpeedAccTradeoff} show that only SSD-like OD can adopt lightweight backbones without a huge drop in accuracy.

One more promising technique to have a lightweight and well-performed solution is knowledge distillation \cite{hinton_dist}. There are plenty of works in this direction for classification task \cite{moonshine}, \cite{do_deep}, \cite{more_attention}. For OD this topic is not so well explored. In \cite{leo_dm} authors adopt Faster R-CNN as an object detection framework to apply distillation and propose a set of steps, which need to be done in order to make distillation work. Despite it looks promisingly, one should have deep teacher model trained first, which requires additional time and sometimes data (to prevent overfitting). Nevertheless, our findings are complementary and can be used along with distillation.

Most of the modern OD is based on backbones pre-trained on ImageNet.
In many cases, pre-training is a separate task which usually requires a lot of time.
Recent works \cite{DSOD}, \cite{GPDSOD} suggest the way how to specifically design CNN, which can be trained directly from scratch for OD. Here we propose steps how to train lightweight OD directly, without specifically designed CNN blocks or need for many hours backbone pre-training on additional data.

Often OD suffers from false positives. No one will deploy OD network in the application if it regularly produces false alarms. One can say, the better detector accuracy, the less number of false positives. But we know, that there is a speed/accuracy trade-off. In the next chapter, we propose a simple method, which allows decreasing the confidence of false positives.

Usually, before running any OD, one should select the threshold of confidence value for a detector. This is the number, above which we consider all detected objects as positives, and the objects below such threshold are considered false positives. So, when running a good OD one will see the box around the object most of the time, but sometimes it blinks. This happens due to the low confidence value of the detected object, so it is filtered by the threshold. It means, that we missed the object in some frames. Such a situation can be compensated with trackers \cite{Mot}. However, a tracker is a separate algorithm, that can be computationally expensive. We outline the extremely cheap tracking strategy, based on re-detection, which utilizes the nature of OD.

\section{Designing Object Detector for Deployment}
In this section we consider all the aspects of designing lightweight object detection architecture for deployment, which is able to run with real-time speed on edge computing devices (at the edge).
Our target use case is OD for ADAS scenario, so the final detector is able to recognize objects of two classes: pedestrian and vehicle (the last includes cars, trucks, buses, etc.). ADAS typically receive an input from a monocular RGB camera and camera is usually mounted inside a car on a windshield or on the top of the car and provides video stream with 16:9 aspect ratio. 
Despite that, all findings and insights can be applied to other classes of objects due to their high-level structure.

%To avoid license issues while open-sourcing the trained model, we collected private train and test base.
In our research we used self-collected datasets to evaluate the accuracy of the final model. They consist of representative sets of objects captured from several cameras under various weather conditions and containing multiple road scenes, like city road, countryside, highway, etc.

\subsection{Design Practices}
\subsubsection{Real-time CNN}
As it was mentioned, there are many object detection frameworks like Faster R-CNN, R-FCN, and SSD. Moreover, many variations of them have been recently designed to improve the quality of the original ones \cite{MSCNN}, \cite{FocalLoss}, \cite{RainbowSSD}. In our work, we chose SSD as a detection architecture based on the comparison made in \cite{SpeedAccTradeoff}. It was shown there, that SSD performs not as well as two-stage detectors like Faster R-CNN or R-FCN in general, but outperforms them with lightweight backbone.
Thus following this fact we used MobileNet \cite{MobileNet} as a feature extractor inside SSD detection framework since it is light in terms of computational complexity as well as in the number of parameters.
Furthermore, we applied several modifications on top of it and inside SSD pipeline to be able to run it in real-time on a mainstream CPU.

\begin{table*}[h!]

%        \footnotesize
\caption{Ablation study of MobileNet+SSD improvements on COCO \texttt{minival} set.}
\label{tab:freq}
\sisetup{
table-number-alignment = center,
}
\begin{tabular*}{\linewidth}{@{\extracolsep{\fill}} ccccccc}
  \toprule
{Improvement} & {} & {} & {} & {}  & {}  & {}\\
  \midrule
{Resolution 300x300}        & * & * &   &   &   &    \\
{Depth-wise head}           &   & * & * &   & * & *  \\
{Resolution 672x384}        &   &   & * & * & * & *  \\
{Two-stage pre-training}    &   &   &   & * & * & *  \\
{Extra predictors}          &   &   &   &   &   & *  \\
 \midrule
   {mAP@0.5IOU} & 0.359 & 0.357 & 0.379 & 0.387 & 0.393 & \textbf{0.411} \\
 \midrule
   {GMAC} & 2.3 & 2.2 & 6.0 & 6.2 & 6.0 & 8.8\\
  \midrule
   {Millions parameters} & 16.4 & 14.8 & 14.8 & 16.4 & 14.8 & 21.1\\
 \bottomrule
\end{tabular*}
\end{table*}

\textit{Resolution.} We used input resolution and aspect ratio different from the original SSD by the following two reasons. One of them is to improve the detection of small objects. 
We increased the input resolution of CNN to 672x384 from the default 300x300. It helps to recognize pedestrians with a minimum size of 40x80 and vehicles with a minimum size of 40x30 on a 720p frame. Another one is that this resolution has the aspect ratio close to 16:9, used in popular image formats like 720p or 1080p. It means that the loss of information along ``width'' dimension is less than in the case of square resolution.

\textit{Depth-wise head.} Besides the backbone, we also used the depth-wise block in SSD ``head''. In the recent MobileNetV2 paper \cite{MobileNetV2} the similar architecture was called SSD-lite. Authors argue that such change reduces computational complexity but does not affect quality dramatically.

\begin{figure}
\centering
\includegraphics[height=5.5cm]{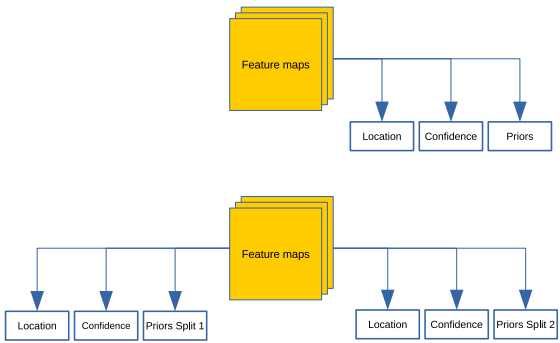}
\caption{Original SSD prediction branch ({\it top}), and split branches ({\it bottom}).}
\label{fig:PriorsSplit}
\end{figure}

\textit{Extended SSD.} We used more prediction branches in SSD to improve handling of small and medium-size objects. We added two additional branches (one for small and one for medium size) and put them to the same feature maps as the first two in the original MobileNet+SSD architecture \cite{MobileNet}. This also forced us to change sizes of prior boxes placed on the same feature maps. For example, if originally prior boxes had parameters $min\_size$, $max\_size$ then they would be evenly split and have parameters $min\_size$, $\frac{min\_size+max\_size}{2}$ and $\frac{min\_size+max\_size}{2}$, $max\_size$ accordingly. Fig.~\ref{fig:PriorsSplit} shows a scheme of such split.

It may seem, that these branches can significantly increase GMAC number and slow down the inference time since they placed on the feature map with the highest spatial resolution. However for the target use case, when we need to detect just two classes, this change is not so dramatic, while allows to reasonably improve quality. Table~\ref{table:extra_branches} shows such comparison for the networks with 672x384 input resolution after two-stage pre-training and depth-wise head.

\setlength{\tabcolsep}{4pt}
\begin{table}
\begin{center}
\caption{Average Precision for person and car classes on the COCO \texttt{minival} set.}
\label{table:extra_branches}
\begin{tabular}{lccc}
\hline 
\multirow{2}{*}{Experiment} & \multicolumn{2}{c}{AP@0.5IOU} & \multirow{2}{*}{GMAC}\\
 & person & car\\
\hline 
Base model & 0.619 & 0.417 & 3.3\\
Model with extra predictors & 0.635 & 0.423 & 3.6\\
\hline
\end{tabular}
\end{center}
\end{table}
\setlength{\tabcolsep}{1.4pt}
Changes from each design choice are summarized in the Table~\ref{tab:freq}.

\textit{Pruning.} Since we are solving two-class detection problem, pedestrians, and vehicles, we can use fewer channels in most layers. It can be done by applying pruning methods which remove the whole convolutional kernels to obtain immediate inference speed improvement. These methods can be based on some straightforward strategy, such as random sampling, or more sophisticated algorithms which consider the importance of filters \cite{LiPruning}, \cite{WeiPruning}. In \cite{towards_lightweight} authors show, that different pruning methods give comparable results for a similar problem. Guided by considerations of simplicity and ease of reproducibility, we used the random filter sampling. After pruning the network, one more training stage for a couple of epochs is performed to adopt the weights, see results in Table~\ref{table:pruning_results}. The pruned model shows slightly better results both for pedestrians and vehicles because pruning has a regularization effect and can help convergence.

\setlength{\tabcolsep}{4pt}
\begin{table}[!ht]
\begin{center}
\caption{Pruning results on private validation set for two-class model.}
\label{table:pruning_results}
\begin{tabular}{lccc}
\hline\noalign{\smallskip}
\multirow{2}{*}{Experiment} & \multicolumn{2}{c}{AP@0.5IOU} & \multirow{2}{*}{GMAC}\\
 & pedestrian & vehicle\\
\noalign{\smallskip}
\hline
\noalign{\smallskip}
Base model & 0.8815 & 0.9069 & 3.6\\
Pruned model & 0.8836 & 0.9071 & 1.96\\
\hline
\end{tabular}
\end{center}
\end{table}
\setlength{\tabcolsep}{1.4pt}

\subsubsection{Two-stage pre-training}
The one important aspect of designing the OD model is that in order to achieve a sufficient quality a backbone should be pre-trained on some diverse dataset, such as ImageNet, which contains millions of images. However, this process might be time-consuming and has some disadvantages, such as learning bias, domain mismatch \cite{DSOD}, etc. While experimenting with various datasets, we found that for object detection use case it is possible to use just COCO \cite{MSCOCO} dataset to get decent OD. %without backbone pre-training on ImageNet.

To train model from scratch a good gradient estimation is required. Thus a large batch is essential in this case. However, since the input resolution of the network was increased, not so many images may fit into memory, the actual number depends on specific device configuration. We can provide a batch size of 96 images during the training, which leads to weak accuracy results.
That is why we propose two-stage pre-training on COCO. At the first stage, we trained MobileNet+SSD on the original resolution of 300x300 pixels with large batch size. After that, we changed the resolution to the target one,  adjusted size of prior boxes and used weights of small resolution model to initialize stage of fine-tuning with smaller batch size. We did not freeze any weights of layers during both stages. The intermediate results for single-stage training, the results of the two-stage training scheme and this scheme with additional prediction layers are shown in the Table~\ref{tab:freq}.

To compare we trained MobileNet+SSD 300x300 on both ImageNet+COCO and with the proposed procedure on COCO only. Table~\ref{table:imagenet_results} shows results of these experiments. Both results are similar, hence such a two-stage scheme allows to use just a single dataset and avoid spending time for additional hyperparameters tuning for backbone pre-training.

\setlength{\tabcolsep}{4pt}
\begin{table}
\begin{center}
\caption{Results on COCO \texttt{minival} set of MobileNet+SSD after pre-training on ImageNet+COCO and only on COCO.}
\label{table:imagenet_results}
\begin{tabular}{lc}
\hline\noalign{\smallskip}
Experiment & mAP@0.5IOU\\
\noalign{\smallskip}
\hline
\noalign{\smallskip}
ImageNet+COCO  & 0.363\\
Two-stage scheme only on COCO  & 0.359\\
\hline
\end{tabular}
\end{center}
\end{table}
\setlength{\tabcolsep}{1.4pt}
 
After two-stage pre-training, we fine-tuned the final topology on the proprietary dataset.

\subsubsection{False positives suppression}

\begin{figure*}
\centering
\includegraphics[height=3.5cm]{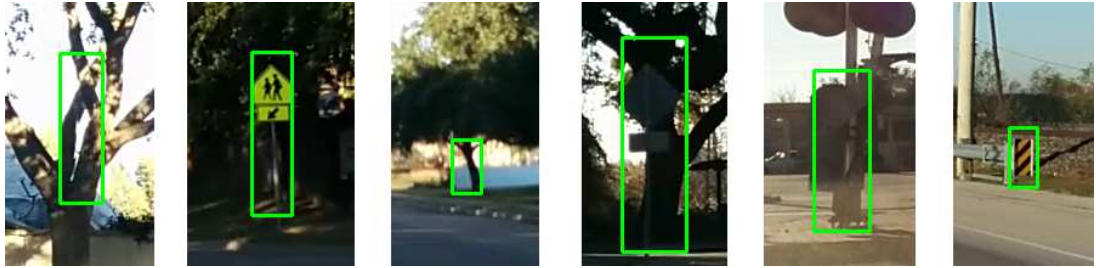}
\caption{Typical False Alarms with high confidence for a pedestrian detector.}
\label{fig:PDFA}
\end{figure*}

Pedestrian detection remains a hard task to solve due to large-scale and appearance variability. In \cite{PDFP} authors show, that objects with significant horizontal gradient, like poles, trees trunks, etc. are strong (have high confidence value) false positives, classified as pedestrians. These objects are typical for the road scenes, see Fig.~\ref{fig:PDFA}, so we set them as the first candidates to suppress.

To address the problem, we collected additional dataset of frames with only such objects, so these images don't contain positives. This is done to balance positives and hard false positives. It is also complicated to find hard false positive and positives in the same image. Furthermore, such images do not require any annotation, thus making the process of gathering them cheap. Our aim is to show, how to use such kind of data to suppress false positives, without reducing in detection quality. So we ran already trained detector on this dataset and left frames only with false positives, which have reasonable confidence (more than 0.3). It was done to make sure that each additional frame, containing false positive, will have a strong impact on training. Then batch size was increased by $\sim$30\% to include such images, so the total size of training dataset increased by $\sim$30\%. By default, SSD framework doesn't compute loss from frames without positives, so we made modifications in the loss layer to allow contribution from such frames and continued training for 5 epochs with the same parameters. This trick leads to the false positive rate reduction, but as well as overall accuracy degradation. We hypothesize that such false positives introduce difficulties into the training process, so network goes fast from local optima and started to learn how to filter false positives, but not how to detect pedestrians well. To remedy the situation, we adjusted the learning rate by decreasing it twice from the default one. This prevents the network to go so far from local optima, while reduces the confidence of false positives.

\setlength{\tabcolsep}{4pt}
\begin{table}
\begin{center}
\caption{Results of false positive suppression scheme.}
\label{table:fpresults}
\begin{tabular}{lc}
\hline\noalign{\smallskip}
Experiment & \makecell[tc]{Miss Rate@0.1 \\ False Positives Per Image}\\
\noalign{\smallskip}
\hline
\noalign{\smallskip}
Base model  & 0.24\\
\makecell[tl]{Model trained with \\ false positives}  & 0.14\\
\hline
\end{tabular}
\end{center}
\end{table}
\setlength{\tabcolsep}{1.4pt}
The results in the Table~\ref{table:fpresults} shown, that using proposed scheme, the detector will find more positives with the same number of false positives. The final average precision of the network remains the same, so this scheme does not actually eliminate strong false positives. However, it reduces their confidence, making them not so strong.

\subsubsection{Results post-processing}
When working with a video stream, not just with a single image, it is important to have an auxiliary part in an overall pipeline, which is responsible for tracking objects, if the main detector fails. There is a lot of research done in this field \cite{median_flow}, \cite{bradski98}, \cite{kcf}, \cite{mdnet}, \cite{sanet}, \cite{eco}. While simple approaches are able to run in real-time, they usually stuck on background objects, if the tracked object was occluded. The more complex solutions can handle this, but they usually require extra computations to be able to discriminate background. We propose tracking strategy, which utilizes the nature of OD, so it distinguishes object versus background and uses almost no extra computations.

Almost all DL-based OD perform the final classification of multiple proposals. Some of them have high confidence, which passes the threshold, the rest are usually discarded. The idea of the proposed tracking method is to match detections with reasonable confidence value from the current frame to detections from the previous frame. As a similarity measure, the popular Intersection Over Union (IOU) metric is used. We match all detections with the confidence higher than 0.2 and choose the best one in IOU metric. So, if for low-confident detection on the current frame there is a match on the previous frame, then this detection is retained despite its low confidence. Such simple re-detection approach allows to compensate detector errors and the negative influence of setting strong detection threshold value, so detections blink less often. Moreover, such retained detections don't stick at the background, because the detector discriminates them and even doesn't give proposals when the object is occluded by something or left the frame. The runtime of this procedure is \textit{less than 0.1 ms} in a challenging road scene with more than 10 objects on our test system, so it is reasonable to use it.

\subsection{Inference with OpenVINO}
One important thing that also should be considered is the inference engine. That is why hardware vendors provide highly optimized inference frameworks such as Nvidia Tensor RT \cite{TensorRT} or Intel OpenVINO \cite{OpenVINO}. 

In our work we used OpenVINO and its Intel DL Deployment Toolkit as a target solution for inference. OpenVINO is able to import models from many DL frameworks and optimize them for various Intel hardware, like CPUs, GPUs, FPGAs or Movidius VPUs.

\subsection{Results}
We performed all the experiments using Caffe framework with additional layers to implement SSD and depth-wise convolutions.

Table~\ref{table:AccuracyResults} shows an accuracy of our final MobileNet+SSD architecture on 672x384 resolution. 
%This network is available as a part of Intel OpenVINO pre-trained models under the name \textbf{``pedestrian-and-vehicle-detector-adas-0001''}.%

\setlength{\tabcolsep}{4pt}
\begin{table}
\begin{center}
\caption{Test results for pedestrians and vehicles.}
\label{table:AccuracyResults}
\begin{tabular}{lcc}
\hline 
Object type
& AP@IOU0.5
& \makecell{Minimum height of \\ stably detected objects \\ on 720p frame}\\\hline
Pedestrian  & 0.88 & 80 pixels\\
Vehicle & 0.906 & 30 pixels\\
\hline
\end{tabular}
\end{center}
\end{table}
\setlength{\tabcolsep}{1.4pt}

%The computational complexity of our final CNN is 2.76 GFLOPs. %
To measure its performance we used publicly available OpenVINO toolkit and ran experiments on Intel Core i5-6500 CPU, which can be used in the edge devices. In the Table~\ref{table:PerformanceResults} the results are also compared with naive inference using the Caffe framework compiled with Intel MKL library. It can be seen, that designed OD can be offloaded to the integrated GPU, while still run in real-time, so CPU will be available for other tasks. Table~\ref{table:TargetPerformanceResults} shows the performance of the designed OD on the target hardware in comparison with the simple two-class baseline network, which is SSD with MobileNet backbone without proposed improvements, on the identical resolution.

\setlength{\tabcolsep}{4pt}
\begin{table}
\begin{center}
\caption{Performance results (Frames Per Second) with OpenVINO on Intel Core i5-6500 CPU@2.90GHz with integrated GPU HD Graphics 530@1.00 GHz.}
\label{table:PerformanceResults}
\begin{tabular}{ccc}
\hline\noalign{\smallskip}
CPU & GPU & Caffe CPU\\
\noalign{\smallskip}
\hline
\noalign{\smallskip}
63.51 & 35.22 & 4.82\\
\hline
\end{tabular}
\end{center}
\end{table}
\setlength{\tabcolsep}{1.4pt}

\setlength{\tabcolsep}{4pt}
\begin{table}
\begin{center}
\caption{Performance comparison for two-class models on Intel Core i5-6500 CPU@2.90GHz.}
\label{table:TargetPerformanceResults}
\begin{tabular}{lc}
\hline\noalign{\smallskip}
Model & FPS\\
\noalign{\smallskip}
\hline
\noalign{\smallskip}
Baseline 672x384 & 40.03\\
Final model & 63.51\\
\hline
\end{tabular}
\end{center}
\end{table}
\setlength{\tabcolsep}{1.4pt}

\begin{figure*}[ht]
\centering
\includegraphics[height=8.5cm]{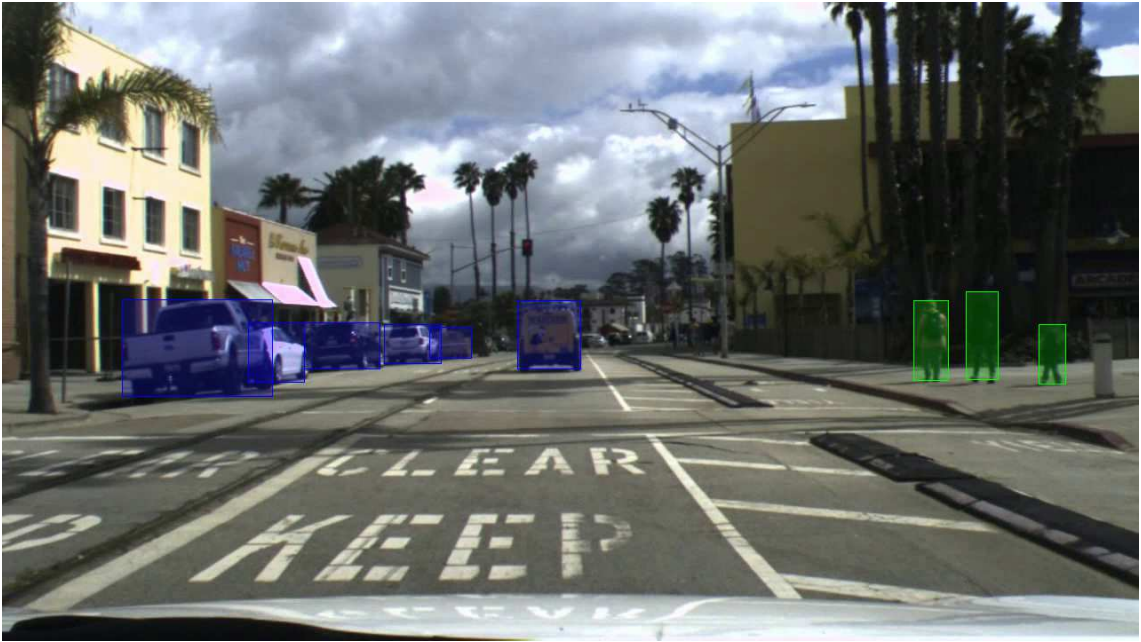}
\caption{Detection example in real road conditions.}
\label{fig:Example}
\end{figure*}

\section{Conclusions}

In this work, we have outlined some important problems of developing DL-based object detector and provide solutions to deal with them. Based on these insights we developed a lightweight CNN which shows 60+ frames per second of inference speed with OpenVINO toolkit on a general purpose CPU. We focused on ADAS case, however, such practices can be applied to other domains. It worth to note, that usually real systems consist of complex pipelines, which combine multiple tasks sharing the same hardware. So every component of such systems should operate faster than real-time, to allow the whole pipeline running in real-time. Thus we believe, that the described practices are important to develop performance-critical OD for applications.%We hope, that described practices are helpful to develop OD for applications.% one can design more complex pipeline, which solves multiple tasks, and deploy it to real system, while be able to operate in real-time.

Using quantization or even binarization of the model weights can further improve the inference speed as well as more sophisticated pruning methods. Moreover, designing CNN in a hardware-friendly way may further boost the performance. We left the evaluation and development of such practices for the future research.

Our final model can be downloaded as a part of \href{https://software.intel.com/en-us/openvino-toolkit/choose-download}{OpenVINO toolkit}. The network description is available in \href{https://github.com/opencv/open_model_zoo/tree/2018/intel_models}{Open Model Zoo} repository under the name \textit{pedestrian-and-vehicle-detector-adas-0001}.

{\small
\bibliographystyle{ieee}
\bibliography{arxiv}
}

\end{document}